\documentclass[sigconf]{acmart}
\AtBeginDocument{%
  \providecommand\BibTeX{{%
    \normalfont B\kern-0.5em{\scshape i\kern-0.25em b}\kern-0.8em\TeX}}}
\setcopyright{acmcopyright}
\copyrightyear{2020}
\acmYear{2020}
\acmDOI{10.1145/1122445.1122456}

\acmConference[Woodstock '20]{Woodstock '20: ACM International Conference On Information And Knowledge Management}{June 03--05, 2018}{Woodstock, NY}
\acmBooktitle{Woodstock '18: ACM Symposium on Neural Gaze Detection,
  June 03--05, 2018, Woodstock, NY}
\acmPrice{15.00}
\acmISBN{978-1-4503-XXXX-X/18/06}

\usepackage{algorithm}
\usepackage{algorithmicx}
\usepackage{algpseudocode}
\usepackage{amsmath}
\begin{document}
\title{Object Hider: Adversarial Patch Attack Against Object Detectors}

\author{Yusheng Zhao}
\authornote{Both authors contributed equally to this research.}
\affiliation{
  \institution{Beihang University}
  \streetaddress{street address}
  \city{Beijing}
  \country{China}}
\email{zhaoyusheng@buaa.edu.cn}

\author{Huanqian Yan}
\authornotemark[1]
\affiliation{
  \institution{Beihang University}
  \streetaddress{street address}
  \city{Beijing}
  \country{China}}
\email{yanhq@buaa.edu.cn}

\author{Xingxing Wei}
\authornote{Corresponding author.}
\affiliation{
  \institution{Beihang University}
  \streetaddress{street address}
  \city{Beijing}
  \country{China}}
\email{xxwei@buaa.edu.cn}

\renewcommand{\shortauthors}{Zhao and Yan, et al.}

\begin{abstract}
Deep neural networks have been widely used in many computer vision tasks. However, it is proved that they are susceptible to small, imperceptible perturbations added to the input. Inputs with elaborately designed perturbations that can fool deep learning models are called adversarial examples, and they have drawn great concerns about the safety of deep neural networks. Object detection algorithms are designed to locate and classify objects in images or videos and they are the core of many computer vision tasks, which have great research value and wide applications. In this paper, we focus on adversarial attack on some state-of-the-art object detection models. As a practical alternative, we use adversarial patches for the attack. Two adversarial patch generation algorithms have been proposed: the heatmap-based algorithm and the consensus-based algorithm. The experiment results have shown that the proposed methods are highly effective, transferable and generic. Additionally, we have applied the proposed methods to competition \textit{Adversarial Challenge on Object Detection} that is organized by Alibaba on the Tianchi platform and won top 7 in 1701 teams. Code is available at \textit{https://github.com/FenHua/DetDak}
\end{abstract}


\maketitle

\begin{figure*}[t]
  \centering
  \includegraphics[width=0.95\textwidth]{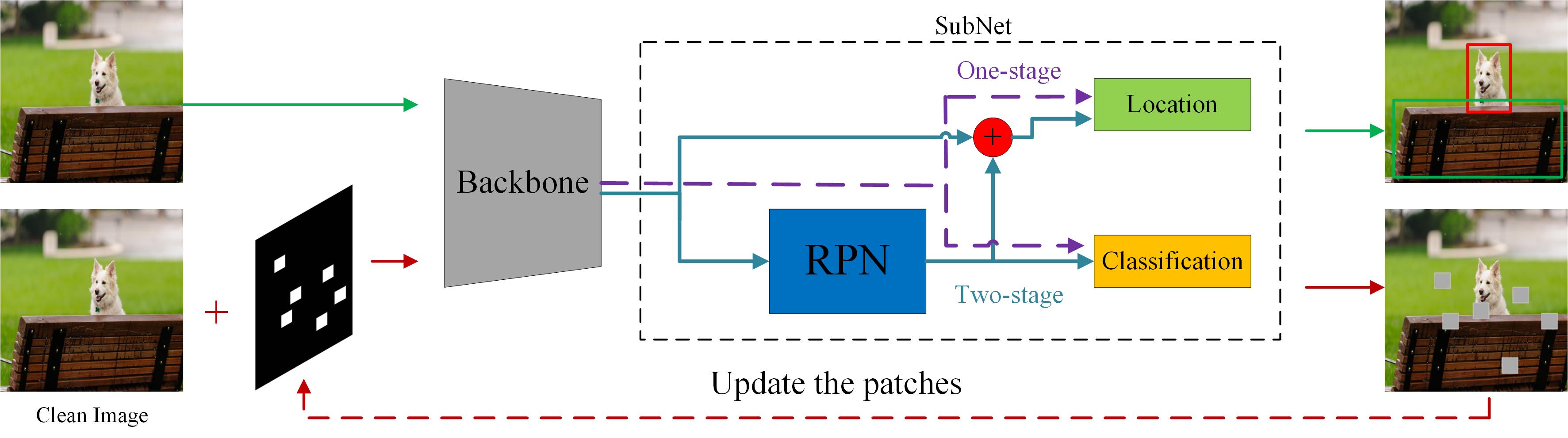}
  \caption{The framework of our attack method. It is an adversarial patch attack algorithm for object detection. The strategies of generating patches are described in Section 2.}
  \Description{description}
  \label{Figure 1}
\end{figure*}

\section{Introduction}
While being widely used in many fields, deep neural networks are shown to be vulnerable to adversarial examples \cite{Szegedy}. Many early studies of adversarial examples focused on the classification task, adding perturbation on the entire image. However, in real world applications like autonomous vehicles and surveillance, such perturbation is hard to implement. Because of this, recent studies focus mainly on adversarial patches, which restrict the perturbation to a small region like a rectangular area. This makes adversarial examples more practical and easier to implement.

Object detection is an important part of computer vision and enables many tasks like autonomous driving, visual question answering and surveillance. However, there are relatively few studies on the adversarial attack of object detection models, especially for the purpose of making the objects disappear. Since object detection models have been used in many life-concerning applications, research about the fragility of these models is of great importance.

Therefore, we aim to investigate the vulnerability of object detection algorithms in this work and attack four state-of-the-art object detection models provided by Alibaba Group on the Tianchi platform, including two white-box models --- YOLOv4 \cite{YOLOv4} and Faster RCNN \cite{FasterRCNN}, and two black-box models to test the transferability of the proposed algorithm. The purpose of the designed methods is to blind the detection models with the restricted patches. The framework of adversarial attacking is shown in Figure \ref{Figure 1}.

We discover that the locations of adversarial patches are crucial to the attack, so we focus on locating the patches and propose two patch selection algorithms: the heatmap-based algorithm and the consensus-based algorithm. The heatmap-based algorithm is an improved version of Grad-CAM \cite{Grad-CAM}, which introduced the idea of heatmap to visualize the gradients of intermediate convolutional layers in image classifiers. We modify and improve the algorithm to make it suitable for visualizing the gradients in object detection models and use the heatmap to select patches. To the best our knowledge, it is the first Grad-CAM-like algorithm designed specifically for the object detection task. The consensus-based algorithm is another novel patch selecting method. It chooses patch locations by attacking several target models and combining the results with a voting strategy, which can make the location of the patch more precise and the adversarial examples more transferable.

We test our attacking algorithm with the proposed patch selection algorithms on the dataset provided by Alibaba Group. The result shows that the proposed algorithms are highly competitive. In brief, the main contributions can be summarized as follows:
\begin{itemize}
    \item We improve the Grad-CAM algorithm to make it more suitable for analysing the gradients of object detection models and use it for the heatmap-based attack.
    \item We propose consensus-based attack algorithm that is very powerful for attacking object detection models.
    \item The experimental results show that the proposed attacking methods are competitive and generic.
\end{itemize}
The rest of this paper is organized as follows. The proposed algorithms are described in Section 2. The experimental results and analysis are presented in Section 3. Finally, we summarize the work in Section 4.
\section{Methods}
Two methods have been designed for generating patches: the heatmap-based algorithm and the consensus-based algorithm. In this Section, two proposed methods are introduced in details. The adversarial attack algorithm with patches is also presented concretely at end of this Section.
\subsection{Heatmap-based Algorithm}
Grad-CAM\cite{Grad-CAM} is a popular tool for visualizing the derivative of the output with respect to an intermediate convolutional layer. It introduced the idea of heatmap --- important regions of the input are hotter in the heatmap. The heatmap is a function of the gradients of the output with respect to an intermediate layer. However, the original Grad-CAM algorithm is designed for classification models and thus cannot be used directly in our task. On the one hand, object detection tasks usually have multiple objects for the input image, while the classification task only have one. On the other hand, the size of different objects could have a significant influence on the heatmap, so we cannot directly add the gradients together when computing the heatmap. 

Therefore, an improved Grad-CAM algorithm is proposed for selecting patches. Firstly, we adopt the element-wise multiplication of the gradients and activations, which preserves spatial information of the gradients and the intermediate layer. Secondly, we normalize the heatmap data of all bounding boxes and combine them together to get the heatmap of the entire image. Thirdly, we use several intermediate layers of the backbone for computing the heatmap, which can combine both lower-level features and higher-level features. Finally, we get the patch mask according to the values of the heatmap.

Mathematically, we calculate the heatmap $H$ using the following formula:
\begin{equation}
H = \sum_{a\in \mathcal A} H_a,
\end{equation}
where $\mathcal A$ is the set of several activation layers (like conv56, conv92 in YOLOv4). $H_a$ is the heatmap of a single activation layer $a$, which is defined as:
\begin{equation}
H_a = \sum_{b\in B} \frac{h^b-\mathrm E[h^b]}{\sqrt{\mathrm {Var}[h^b]}} \cdot \sqrt{Area_b},
\end{equation}
where $h^b$ represents the heatmap of a single bounding box $b$, the mean $\mathrm E(h^b)$ and the variance $\mathrm{Var}(h^b)$ are used for normalization. Besides, $Area_b$ is the area of the bounding box $b$ and $\sqrt{Area_b}$ is used in the normalization to avoid small bounding boxes from being too dominant. We compute $h^b$ as
\begin{equation}
h^b = \max(0, \sum_k \frac {\partial y^b} {\partial A_{ij}^k} \odot A_{ij}^k),
\end{equation}
where $A_{ij}^k$ denotes the activation of a convolutional layer at channel $k$ and location $(i,j)$, $\odot$ represents the element-wise product and $y^b$ is the highest confidence score of the bounding box $b$.

Finally, we use some Gaussian filters to post-process the heatmap to make it more smooth. Combining the heatmaps of several object detection models, we can choose the patches in hot regions of the input image.
\subsection{Consensus-based Algorithm}
Although the heatmap-based algorithm exploits the gradient information of the models, it is separated from the attacking process. Besides, we find that the sensitive locations of the input image might change over time when attacking algorithms are performed iteratively. Therefore, we propose another method for patch selection: the consensus-based algorithm.

First of all, we perform the Fast Gradient Sign Method iteratively with $L2$-norm regularization on the target models respectively. Our loss function $J_{L2}$ is originally defined as:
\begin{align}
    & J_{model}=\sum_{i\in \{j|s_j>t\}} s_i,\\
    & J_{L2} = J_{model} + \omega\cdot ||P||_2^2,
\end{align}
where $s_i$ is the confidence score of each bounding box of the corresponding model, $t$ is the confidence threshold (we use 0.3 in our task). Usually, when $s_i>t$, it indicates the bounding box is correct and will appear in the results. So the lower the confidence score $s_i$, the fewer objects can be detected. $P$ represents the perturbation and $\omega$ is a hyper parameter.

In the experiments, we find that the noise perturbations of some models like Faster RCNN are not concentrated, which makes it hard to fuse multiple results. To solve this problem, we modified the loss function of those models:
\begin{equation}
    J_{rcnn}=\gamma\sum s_{key} + (1-\gamma)\sum s_{other},
\end{equation}
where $s_{key}$ is the confidence score of bounding boxes appeared in the clean image and $s_{other}$ is the confidence score of others that do not contain any true objects during the attack. The $\gamma$ is a hyper parameter which is set to 0.9 in our experiments. Such modification could force the perturbation to concentrate on the main objects of the image.

\begin{figure}[t]
  \centering
  \includegraphics[width=0.95\linewidth]{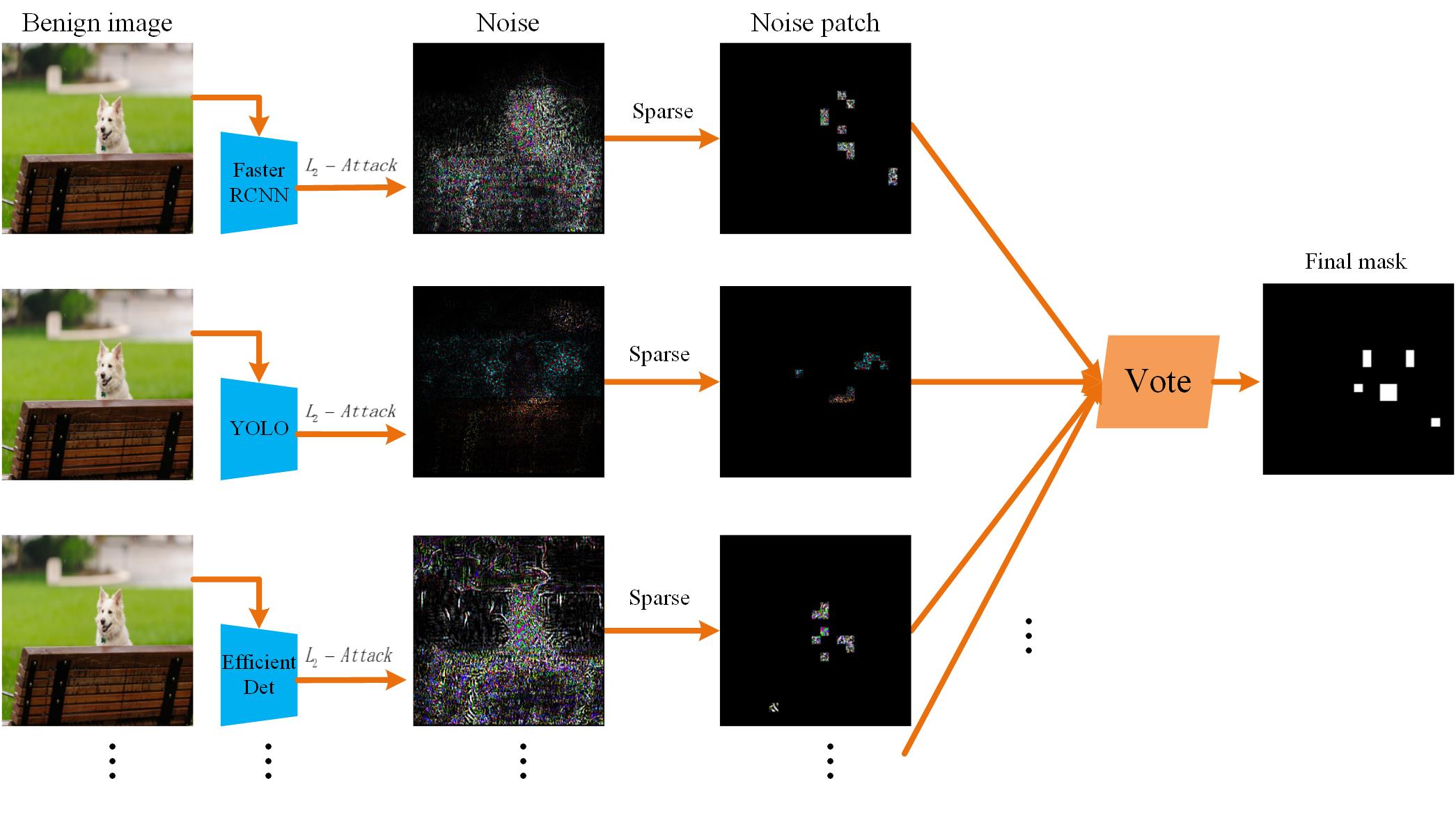}
  \caption{Consensus-based Algorithm which uses a voting method for generating patch masks.}
  \Description{description}
  \label{Figure 2}
\end{figure}

After $L_2$ attack, we can get the noise of the input image of each model. However, we do not mix those noise directly, because they are different in magnitude and it is not easy to balance them. So we sparsify the noise into $n$ patches with a specified scale $\mathcal S$. Next, we take a vote to decide which patch mask should be preserved and which should be discarded. Usually, the greater the perturbation, the more likely it is to be selected as patch. The voting strategy on those noise patches is very helpful for improving performance. The flow of the algorithm is described in Figure \ref{Figure 2}. Here, we introduce EfficientDet \cite{EfficientDet} to join the vote. The more the detection models, the more accurate the voting results and the higher the adaptability. Furthermore, the voting strategy can also improve the transferability of our adversarial patches and the robustness of our algorithm. Additionally, the number of $L_2$ attack iterations is not very sensitive. Even with only 5 iterations, the voting result is still quite decent.
\subsection{Adversarial attack with patches}
After patch mask generation, Fast Gradient Sign Method (FGSM) is used to finish attacking:
\begin{equation}
    \delta := \mathrm{clip}_{[0,255]}(\delta + \alpha \cdot \mathrm{sign}(\nabla_\delta L)),
\end{equation}
where $\delta$ is the parameter in the adversarial patches and $\alpha$ is the learning rate which we refer to \cite{Szegedy} for setting its value. $L$ is the loss function defined as:
\begin{equation}
    L=\sum_{m\in \mathcal M}\sum_{i\in \{j|s_j^{(m)}>t\}} s_i^{(m)},
\end{equation}
where $s_i^{(m)}$ is the confidence score of the $i$-th bounding box of model $m$ and $\mathcal M$ is set of detection models. The loss function $L$ is simple but efficient. Figure \ref{Figure 1} offers a comprehensible description of the attacking algorithm. The detail of the algorithm with consensus-based patch selection algorithm is also described in Algorithm \ref{Algorithm 1}.
\renewcommand{\algorithmicrequire}{\textbf{Input:}}
\renewcommand{\algorithmicensure}{\textbf{Output:}}
\begin{algorithm}[t]
\caption{Consensus-based Attack Algorithm}
\begin{algorithmic}[1]
\Require a clean image $I$, the patch number $n$, the scale of square patches $\mathcal S$, the iteration number for attacking $it$
\Ensure an adversarial image $I^\prime$
\State Let the set of object detection models be $\mathcal M$
\For{model $m \in \mathcal M$}
\State L2 attack on $m$ and get perturbation $P^{(m)}$
\State Get top $n$ noise patches in $P^{(m)}$ as $P_n^{(m)}$ with scale $\mathcal S$
\State Normalize $P_n^{(m)}$
\EndFor
\State $P_n\gets \sum_{m\in \mathcal M} P_n^{(m)}$
\State $\mathcal P\gets$ select top $n$ patch masks in $P_n$ according to the magnitude of the perturbation.
\Repeat
\State Perform FGSM attack on $I$ with patch masks in $\mathcal P$
\State Update the polluted image $I^\prime$.
\State $it\gets it-1$
\Until{$L=0$ or $it=0$} // $L$ is the loss function
\State \Return $I^\prime$
\end{algorithmic}
\label{Algorithm 1}
\end{algorithm}
\section{Experiment}
\begin{figure}[t]
  \centering
  \includegraphics[width=0.98\linewidth]{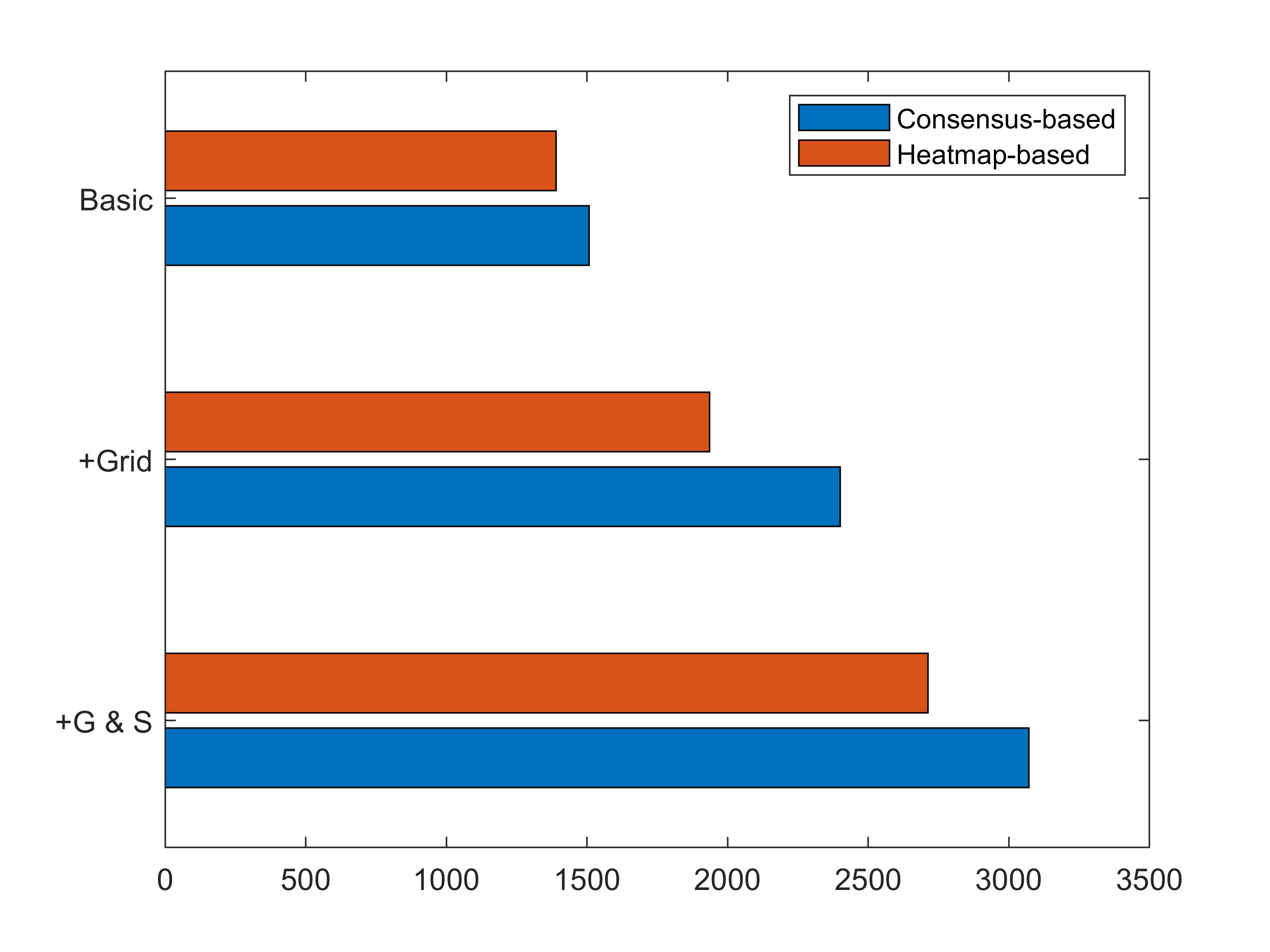}
  \caption{The scores of the two proposed algorithms. "Basic" represents results of two patch generation methods. "+Grid" represents the result of the proposed methods with grid-like patches ensemble. "+G\&$\mathcal S$" means the results of our methods with grid-like and different scale patches ensemble.}
  \Description{description}
  \label{Figure 3}
\end{figure}

We used the proposed methods in \textit{AIC Phase IV CIKM-2020: Adversarial Challenge on Object Detection} competition. The results of two basic proposed methods without any ensemble operations are recorded in Table \ref{tab:result} and shown in Figure \ref{Figure 3}. Even without ensemble operations, the algorithms are quite competitive. In order to reduce the number of pixels of our patches, grid-like patches are designed. 

To get grid-like patches, we performed a element-wise dot product between patch mask $M$ and a grid matrix $G_{ratio}$. The grid-like mask $M^\prime$ can be calculated by:
\begin{equation}
    M^\prime = M \odot G_{ratio},
\end{equation}
where $ratio$ represents the degree of sparsity, and the larger $ratio$ is, the less pixels are used.

Ensemble operation is a common practice in machine learning and we used this in the task. To combine different results for getting better results, an indicator is defined,
\begin{equation}
    \mathrm{FinalScore} = \sum_{i=1}^2\sum_x S(x,x^*,m_i),
\end{equation}
because there is two white-box detector, YOLO and Faster RCNN, can be used, the indicator is the sum of two score functions. $S$ is provided by Alibaba Group:
\begin{equation}
    S(x,x^*,m_i) = (2-\frac {\sum_k R_k}{5000})\cdot \Big(1-\frac{\min\big(BB(x;m_i),BB(x^*;m_i)\big)}{BB(x;m_i)}\Big).
\end{equation}
where $R_i$ is the number of pixels of the $i$-th patch, $x$ is the clean image, $x^*$ is the adversarial example, $m_i$ denotes the $i$-th model, and $BB(x;m_i)$ is the number of bounding boxes detected by $m_i$ on image $x$.

When we perform ensemble operations with grid-like patches through $ratio \in \lbrace 0.5, 0.6, 0.7\rbrace$, we have got more than 2000 score. The results are also recorded in Table \ref{tab:result} and shown in Figure \ref{Figure 3}. As you can see, the effect is obvious.

\begin{table}[t]
\centering
\caption{The results of the proposed two methods.}
\setlength{\tabcolsep}{5mm}{
\begin{tabular}{l|c|c}
\hline
& heatmap-based & Consensus-based \\ \hline
Basic  & 1390 & 1507 \\ \hline
+G  & 1936& 2400 \\ \hline
+G \&  $\mathcal S$ & 2713  & 3071 \\ \hline
\end{tabular}
}
\label{tab:result}
\end{table}
Since we can successfully attack most object detection models with grid-like sparse patches, we can also expand the region of the original patches and sparsify them to cover a larger area while altering a moderate amount of pixels. Besides, we observe that the patches of a fixed size might only be suitable for some images, so we performed an ensemble over patches of different sizes $\mathcal S$. Combining grid operations and the ensemble over patches of three different sizes $\mathcal S$ ($\mathcal S \in \lbrace 20, 50, 70\rbrace$ in our experiments), our best result is over 3000 score.

\begin{figure}[t]
  \centering
  \includegraphics[width=0.95\linewidth]{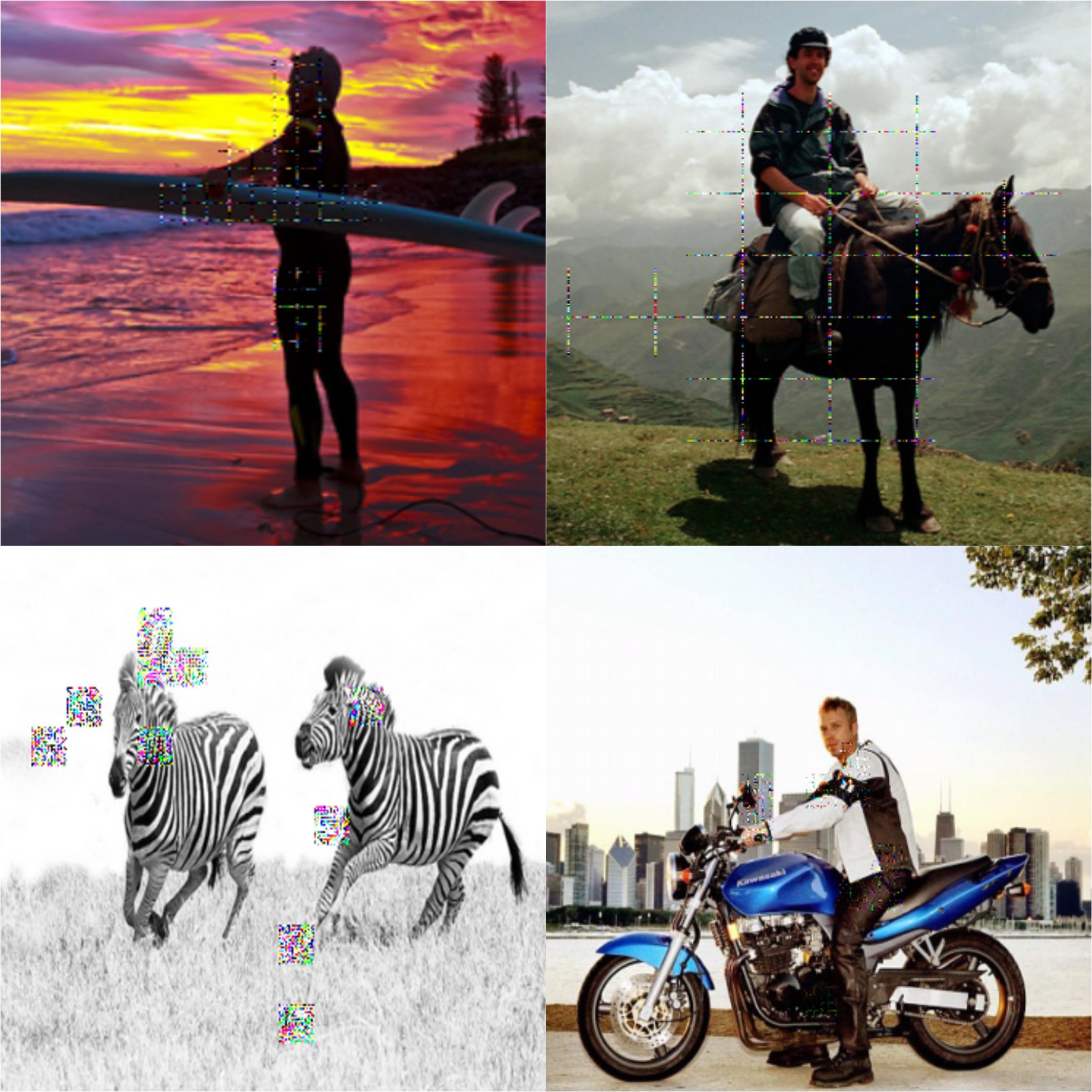}
  \caption{Some adversarial images using the consensus-based attack algorithm with grid-like patches.}
  \Description{description}
  \label{Figure 4}
\end{figure}

Note that the consensus-based algorithm is generally better than the heatmap-based algorithm in the experiments. A possible explanation for this is that the consensus-based algorithm might better combine the target models with the voting process and incorporate the attacking process with the selection process. Some adversarial images are shown in Figure \ref{Figure 4}. As shown, the patches are gridded and have different scales. Since there is no limit to the perturbations, the noise is obvious. In general, the greater the noise is, the better the attack transferibility is.
\section{Conclusion}
In this paper, two adversarial patch generation algorithms have been proposed: heatmap-based and consensus-based patch generation algorithms. The generated patches are efficient and precise. Additionally, they only rely on few pixels but are generic. Furthermore, the proposed attacking methods can misguide state-of-the-art object detection models from detecting the objects. Those adversarial examples are a great threat to deep neural networks deployed in real world applications. Through the study of adversarial examples, the mechanism of deep learning models can be further understood, and robust algorithms can also be proposed. In the future, we will explore how to improve the robustness of current detection models to deal with adversarial examples.

\bibliographystyle{ACM-Reference-Format}
\bibliography{ref}
\end{document}